\documentclass[runningheads]{llncs}

\usepackage{eccv}

\usepackage{eccvabbrv}

\usepackage{graphicx}
\usepackage{booktabs}
\usepackage{algorithm}
\usepackage{algpseudocode}
\floatstyle{boxed}
\restylefloat{algorithm}
\usepackage{multirow}
\usepackage{siunitx}
\usepackage{amsmath}
\usepackage{amssymb}
\usepackage{wrapfig}

\usepackage[accsupp]{axessibility}  

\usepackage[pagebackref,breaklinks,colorlinks,citecolor=eccvblue]{hyperref}
\usepackage{hyperref}
\usepackage{xcolor}

\usepackage{orcidlink}

\begin{document}

\title{Less Tokens, Better Forecasts: Sparse Residual Routing for Efficient Weather Prediction}

\titlerunning{Sparse Residual Routing for Weather Prediction}

\author{Janet Wang\inst{1} \and
Yunbei Zhang\inst{1} \and Lin Zhao\inst{2} \and
Xi Xiao\inst{3} \and Jihun Hamm\inst{1} \and Xiao Wang\inst{4}}

\authorrunning{J.~Wang et al.}

\institute{Tulane University \and
Northeastern University \and University of Alabama at Birmingham \and Oak Ridge National Laboratory
}

\maketitle

\begin{abstract}
Existing ViT-based weather forecasting models apply uniform computation across all spatial tokens, even though nearby atmospheric grid points often contain similar values and large regions evolve smoothly over time. This makes much of the intermediate per-token computation redundant. Standard token-efficiency methods, such as pruning or merging, reduce cost by removing or fusing tokens. However, weather forecasting is a spatiotemporal dense prediction problem in which a history of atmospheric states must be mapped to future values on the original latitude-longitude grid. Thus, every grid cell must retain a physically meaningful representation, especially under autoregressive rollout. We introduce \textbf{Sparse-Reslim}, a parameter-free plug-in routing module that makes sparse token processing compatible with this fixed-grid requirement. Sparse-Reslim routes only 25\% of spatial tokens through the expensive middle transformer blocks and treats those blocks as residual updates: it computes the change produced for the routed tokens and scatters only this delta back to the full sequence. Unselected tokens keep their pre-routing representations exactly, so no grid cell is dropped or replaced by a mask token, and no fusion layer or additional parameters are introduced. Across ERA5 resolutions up to the operational 0.25\textdegree{} standard and two model families, a deterministic Transformer and a diffusion model, Sparse-Reslim improves forecast accuracy on every evaluated variable while substantially reducing cost: training is about $2.5\times$ faster in the main settings and reaches $3.18\times$ speedup at 0.25\textdegree{}, with over $2.2\times$ lower peak memory. A controlled decomposition shows that the accuracy gain comes primarily from sparse routing itself, while random token selection provides an additional regularization benefit without selector overhead. Code is available at \url{https://github.com/janet-sw/Sparse-Reslim}.

\end{abstract}


\section{Introduction}
\label{sec:intro}

Data-driven weather forecasting has reached a turning point \cite{weyn2019can, scher2018toward, dueben2018challenges}. Models trained on reanalysis data now match or exceed the accuracy of operational numerical weather prediction (NWP) systems that took decades to develop, and they do so orders of magnitude faster at inference time~\cite{bi2023accurate, lam2023learning, bodnar2025foundation}. These models operate on high-dimensional gridded atmospheric fields (ERA5 at 0.25\textdegree{} comprises over one million grid points per variable) and must produce spatially dense outputs where every grid cell carries a physically meaningful forecast. As the community pushes toward higher resolutions and larger model capacities, training cost becomes the primary bottleneck, because a finer grid produces more spatial tokens and self-attention scales quadratically with token count, while larger models add further compute and activation memory per token.

A striking pattern across the leading weather models (Pangu-Weather~\cite{bi2023accurate}, ClimaX~\cite{nguyen2023climax}, Stormer~\cite{nguyen2024scaling}, FuXi~\cite{chen2023fuxi}, Aurora~\cite{bodnar2025foundation}, and their generative counterparts GenCast~\cite{price2023gencast} and CorrDiff~\cite{mardani2025residual}) is that they all rely on the Vision Transformer (ViT) or its variants. Self-attention treats every spatial token identically, spending the same $O(L^2 D)$ computation on a token over calm open ocean as on one at the core of an extratropical cyclone. Yet atmospheric fields are highly autocorrelated: large regions of the globe carry smooth, slowly varying conditions whose intermediate representations are largely redundant. This suggests that substantial computation can be saved by processing only a representative subset of tokens in the most expensive layers.

Token pruning~\cite{rao2021dynamicvit, liang2022not, kong2022spvit, xu2022evo}, token merging~\cite{bolya2022token}, and window-based attention~\cite{liu2021swin} have been widely explored for ViT efficiency in the vision community. More recently, Sprint~\cite{park2025sprint} proposed a dense-sparse-dense block schedule for diffusion transformers that routes only a random subset of tokens through the middle blocks, achieving large training speedups for image generation. Directly applying these methods to weather forecasting is nontrivial because weather prediction is a spatiotemporal dense prediction problem: the model conditions on past atmospheric states and must produce a physically meaningful future value at every grid point, often for autoregressive rollout. Token pruning and merging can remove or alter spatial positions, while Sprint replaces skipped tokens with learned mask embeddings and later fuses them through a linear projection. This is acceptable for static image generation, but in weather forecasting it creates a mismatch: late blocks operate on a mixture of true atmospheric representations and synthetic placeholder tokens at fixed grid positions, without explicitly preserving the temporal conditioning structure of the forecast problem.

In this work, we propose \textbf{Sparse-Reslim}, a parameter-free plug-in routing module for ViT-based weather forecasters that makes sparse token processing compatible with dense prediction. Sparse-Reslim partitions the transformer blocks into dense early, sparse middle, and dense late stages. At the entrance to the sparse middle stage, it randomly selects a fraction $r$ of the $L$ spatial tokens, yielding $K=\lfloor rL\rfloor$ routed tokens, and processes only this subset through the expensive middle blocks. Crucially, instead of replacing the full sequence with processed tokens and placeholders, Sparse-Reslim computes the \emph{residual delta} produced by the sparse middle blocks and scatters only this delta back to the original token positions. Unselected tokens receive zero delta and therefore retain their pre-routing representations exactly, ensuring that all $L$ spatial token positions remain valid throughout the model. The module requires no mask embeddings, fusion layers, or additional learnable parameters, and can be enabled in an existing ViT block loop with a single configuration flag.

We validate Sparse-Reslim on ERA5~\cite{hersbach2020era5} at multiple resolutions and on two architecture families: (1)~a deterministic Res-Slim-ViT adapted from the ORBIT-2 exascale system~\cite{wang2025orbit}, and (2)~a diffusion-based generative model following the EDM framework~\cite{karras2022elucidating}. For the generative variant, we use an asymmetric attention design in which self-attention is sparsified while cross-attention retains the full conditioning sequence. Across all settings, a keep ratio of $r{=}0.25$ yields ${\sim}2.5\times$ training speedup. Notably, Sparse-Reslim also improves forecast accuracy over the dense baseline across all evaluated variables and metrics. Our ablations show that this gain is not explained by stochastic regularization alone: deterministic sparse routing already captures most of the improvement, while random token selection provides a smaller additional benefit without selector parameters or scoring overhead. Our main contributions are:

\begin{itemize}
    \item We propose Sparse-Reslim, a sparse routing module for spatiotemporal weather forecasting that preserves all spatial token positions while respecting temporal conditioning. It adds no learnable components and integrates into ViT-based weather forecasters as a plug-in module.
    \item We provide an empirical study spanning two architecture families (deterministic and generative), three ERA5 grid resolutions, and four atmospheric variables. Sparse-Reslim consistently improves accuracy across evaluated settings, and its benefits become more pronounced at higher resolution.
    \item We show that sparse routing is orthogonal to other efficiency strategies, including mixed-precision training and resolution downsampling, and can be combined with them for further gains. To our knowledge, this is the first application of sparse token routing to efficient training of ViT-based weather models in both deterministic and generative settings.
\end{itemize}

\begin{figure}[t]
    \centering
    \includegraphics[width=\textwidth]{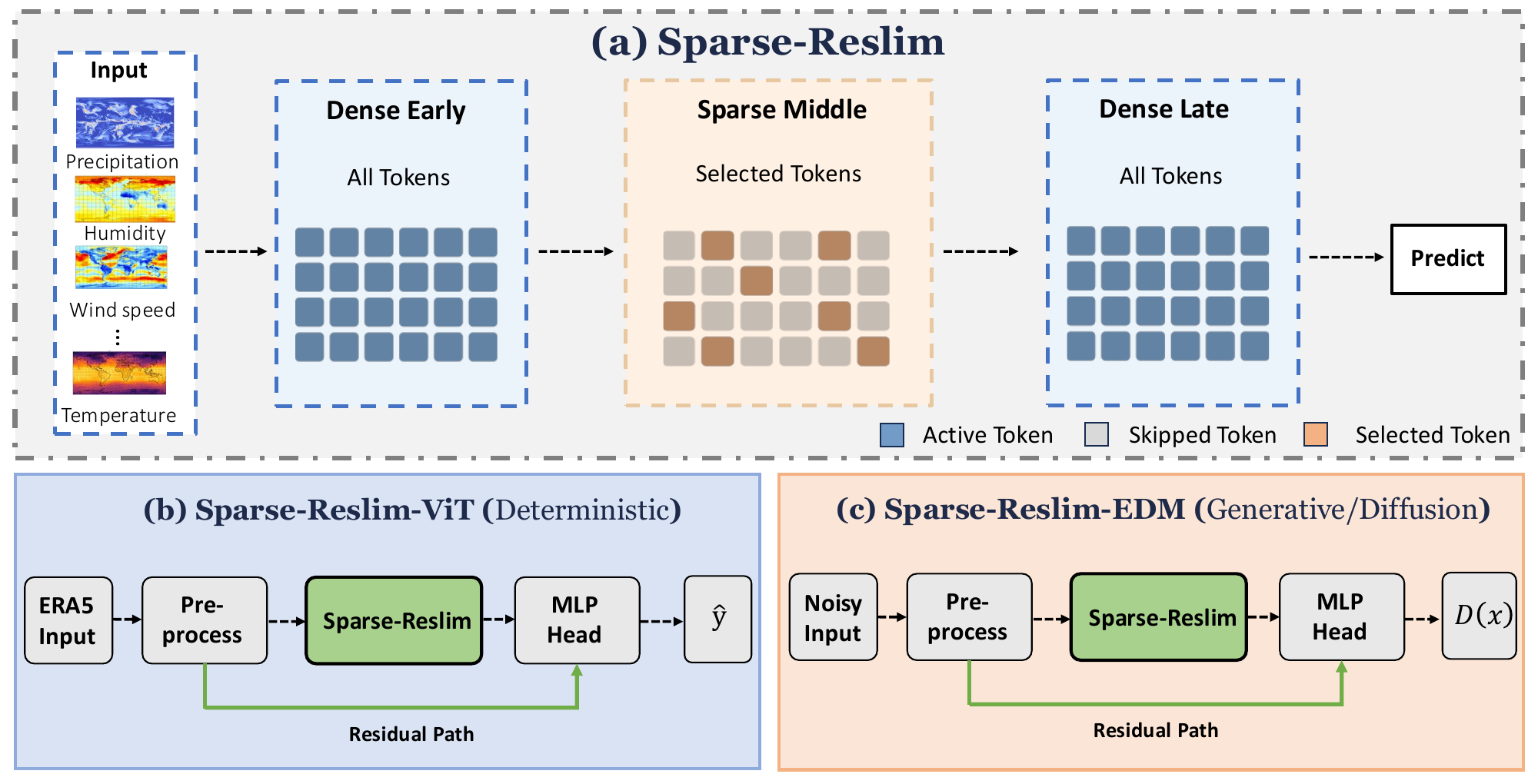}
    \caption{Overview of Sparse-Reslim. (a) Three-stage token routing: dense early ($n_1$ blocks), sparse middle ($n_2$ blocks on $K{=}\lfloor rL\rfloor$ tokens), and dense late ($n_3$ blocks), with unselected tokens following an identity path. (b, c) Integration into the deterministic Res-Slim-ViT and generative Res-Slim-EDM backbones.}
    \label{fig:overview}
\end{figure}

\section{Related Work}
\label{sec:related}

\subsection{AI Weather Forecasting Models}
Data-driven weather forecasting has emerged as a compelling alternative to traditional NWP, reaching comparable or better accuracy at a fraction of the computational cost. The task is to forecast future atmospheric states $\mathbf{X}_{T} \in \mathbb{R}^{V \times H \times W}$ from initial conditions $\mathbf{X}_{0}$, where $V$ denotes physical variables, $H \times W$ the spatial grid, and $T$ the target lead time. Among deterministic approaches, Pangu-Weather~\cite{bi2023accurate} introduced a 3D Earth-Specific Transformer operating on pressure-level volumes, while GraphCast~\cite{lam2023learning} adopted a mesh-based graph neural network for message passing on the sphere. ClimaX~\cite{nguyen2023climax} proposed a variable-agnostic ViT trained across heterogeneous climate datasets, and Stormer~\cite{nguyen2024scaling} combined shifted window attention with weather-specific inductive biases. FengWu~\cite{chen2023fengwu} uses a multi-modal, multi-task cross-modal fusion transformer to push skillful forecasts beyond 10 lead days, and FuXi~\cite{chen2023fuxi} and Aurora~\cite{bodnar2025foundation} extended the frontier with cascade and foundation model designs, respectively.

On the probabilistic side, GenCast~\cite{price2023gencast} applied diffusion models to produce calibrated ensemble forecasts, while OmniCast~\cite{nguyen2025omnicast} and CorrDiff~\cite{mardani2025residual} explored generative approaches for modeling the full distribution of future weather states, addressing the blurring and error accumulation that affect deterministic autoregressive rollout. Despite their architectural diversity, all these models share one trait: they apply uniform computation across all spatial tokens, regardless of local informational complexity. This becomes increasingly wasteful at higher resolutions, where quadratic attention cost dominates training time yet much of the grid is smooth and redundant. Res-Slim-ViT, introduced as part of the ORBIT-2 exascale climate downscaling system~\cite{wang2025orbit}, showed strong scalability through a dual-path design combining a ViT backbone with a convolutional residual path, making it a natural testbed for efficient token routing.

\subsection{Efficient Vision Transformers}
The quadratic cost of self-attention has motivated extensive work on reducing the computational burden of ViTs, and decreasing the number of tokens processed is among the most studied strategies. Attention-based token pruning removes tokens with low attention scores ~\cite{liang2022not, kong2022spvit, xu2022evo, fayyaz2022adaptive}. Several methods go further by training scoring modules that learn task-dependent pruning criteria~\cite{rao2021dynamicvit, rao2023dynamic}. Token merging offers a complementary direction: ToMe~\cite{bolya2022token} condenses similar tokens via bipartite matching, reducing sequence length while preserving representational content. Other approaches operate at the attention level itself, including window-based attention~\cite{liu2021swin} and learned sparse attention patterns~\cite{wei2023sparsifiner, chen2021chasing}.

A more recent line of work targets generative model training specifically. Sprint~\cite{park2025sprint} introduced a dense-sparse-dense block schedule for diffusion transformers: only a fraction of randomly selected tokens pass through the middle blocks, while dropped tokens are replaced by learnable mask embeddings and later fused with shallow dense features via a linear projection. While effective for image generation, Sprint is not directly suited to weather forecasting. Weather models condition on past atmospheric states and must return a complete future state on the original grid. In autoregressive use, this output becomes the next input. Replacing unprocessed positions with mask embeddings therefore does more than add a small projection layer: it breaks the continuity of the atmospheric state representation at fixed grid locations. Our residual delta formulation addresses this gap by keeping unselected tokens as their actual pre-routing representations and scattering back only the updates computed for routed tokens. In the generative setting, we further preserve temporal conditioning by sparsifying self-attention over target tokens while retaining full cross-attention to the conditioning sequence.

\section{Methods}
\label{sec:methods}

This section describes Sparse-Reslim, a drop-in sparse token routing module for ViT-based weather models. Figure~\ref{fig:overview} provides a visual overview of Sparse-Reslim. We first cover the shared tokenization and base architectures (Sec.~\ref{sec:prelim}), then the routing module (Sec.~\ref{sec:sparse_routing}), and its integration into deterministic (Sec.~\ref{sec:deterministic}) and generative (Sec.~\ref{sec:generative}) forecasting pipelines.

\subsection{Preliminaries}
\label{sec:prelim}

\textbf{Tokenization.}
Both architectures share a per-variable tokenization scheme. Each variable $v$ is independently patchified by a dedicated convolutional embedding $\mathtt{PatchEmbed}_v$, producing tokens of dimension $D$. A learnable variable embedding $\mathbf{e}_v$ is added, and variables are aggregated via cross-attention with a shared query $\mathbf{q}\in\mathbb{R}^{D}$:
\begin{equation}
    \mathbf{z}_i = \text{CrossAttn}\!\bigl(\mathbf{q},\; \{\mathbf{p}_{i,v} + \mathbf{e}_v\}_{v\in V}\bigr) \in \mathbb{R}^{D},
    \label{eq:var_agg}
\end{equation}
where $\mathbf{p}_{i,v}$ is the patch token at spatial position $i$ for variable $v$. This yields a sequence $\mathbf{Z} \in \mathbb{R}^{L \times D}$ with $L = (h/p)(w/p)$ tokens, augmented with sinusoidal positional and spatial resolution embeddings.

\textbf{Deterministic forecasting.}
The deterministic model $f_\theta$ directly regresses the forecast target. Given input history $\mathbf{X}$, the history dimension is flattened into the channel dimension and the model predicts:
\begin{equation}
    \hat{Y} = f_\theta(\mathbf{X}) = \text{CNN}_{\text{skip}}(\mathbf{X}) + \text{Unpatchify}\bigl(\text{MLP}_{\text{head}}(\text{ViT}(\mathbf{Z}))\bigr),
    \label{eq:det_forward}
\end{equation}
where $\text{ViT}(\cdot)$ denotes $N$ transformer blocks and $\text{CNN}_{\text{skip}}$ is a convolutional residual path (Conv $\to$ PixelShuffle $\to$ Conv) that bypasses the transformer. The dual-path design lets the lightweight CNN capture dense local patterns, freeing the transformer to focus on long-range dependencies. Training minimizes a latitude-weighted mean squared error:
\begin{equation}
    \mathcal{L}_{\text{det}} = \frac{1}{|V_{\text{out}}|}\sum_{v}\omega_v\cdot \frac{\sum_{i} \cos(\phi_i)\bigl(\hat{Y}_{v,i} - Y_{v,i}\bigr)^{2}}{\sum_{i}\cos(\phi_i)},
    \label{eq:loss_det}
\end{equation}
where $\omega_v$ are per-variable weights and $\phi_i$ is the latitude at grid point $i$.

\textbf{Generative forecasting.}
The diffusion model follows the EDM framework~\cite{karras2022elucidating}, parameterizing a denoiser $D_\theta$ via preconditioning scalars derived from the noise level $\sigma$:
\begin{equation}
    D_\theta(\mathbf{y};\sigma,\mathbf{c}) = c_{\text{skip}}(\sigma)\cdot\mathbf{y} + c_{\text{out}}(\sigma)\cdot F_\theta\!\bigl(c_{\text{in}}(\sigma)\cdot\mathbf{y};\;\sigma,\;\mathbf{c}\bigr),
    \label{eq:edm_precond}
\end{equation}
where $\mathbf{y} = Y + \sigma\boldsymbol{\epsilon}$ is the noisy target with $\boldsymbol{\epsilon}\sim\mathcal{N}(\mathbf{0},\mathbf{I})$, $\mathbf{c}$ represents the conditioning fields (history frames), and $c_{\text{skip}},c_{\text{out}},c_{\text{in}}$ are analytic functions of $\sigma$ and the data standard deviation $\sigma_{\text{data}}$. The raw network $F_\theta$ is a ViT whose blocks contain noise-conditioned FiLM modulation, noise-gated cross-attention, and a CNN skip path:
\begin{equation}
    F_\theta(\cdot) = \text{Unpatchify}\bigl(\text{MLP}_{\text{head}}(\text{ViT}_{\text{edm}}(\mathbf{Z}_x,\mathbf{Z}_c,\mathbf{e}_\sigma))\bigr) + \text{CNN}_{\text{skip}}(c_{\text{in}}\cdot\mathbf{y}).
    \label{eq:edm_Fx}
\end{equation}
Here $\mathbf{Z}_x$ and $\mathbf{Z}_c$ are token sequences from the noisy target and conditions respectively, and $\mathbf{e}_\sigma = \text{MLP}(\ln\sigma/4)$ is the noise embedding. Each EDM block consists of FiLM modulation (\emph{scale}, \emph{shift} from $\mathbf{e}_\sigma$), self-attention, $\sigma$-gated cross-attention ($\text{gate} = \text{sigmoid}(\mathbf{W}\mathbf{e}_\sigma)$), and MLP. Training minimizes the denoising loss:
\begin{equation}
    \mathcal{L}_{\text{edm}} = \mathbb{E}_{\sigma\sim p(\sigma)}\Bigl[\lambda(\sigma)\bigl\|D_\theta(\mathbf{y};\sigma,\mathbf{c}) - Y\bigr\|^2\Bigr],
    \label{eq:loss_edm}
\end{equation}
where $\ln\sigma\sim\mathcal{N}(P_{\text{mean}},P_{\text{std}}^2)$ and $\lambda(\sigma) = (\sigma^2+\sigma_{\text{data}}^2)/(\sigma\cdot\sigma_{\text{data}})^2$ is the weighting.

\subsection{Sparse-Reslim: Sparse Token Routing}
\label{sec:sparse_routing}

\textbf{Motivation.}
Self-attention costs $O(L^2 D)$ per block. For ERA5 at 1.0\textdegree{} with patch size 2, $L{=}16{,}200$ tokens, and the $N{=}12$ transformer blocks dominate training time. Atmospheric fields, however, exhibit strong spatial autocorrelation: nearby grid points carry highly redundant information, especially in the smooth interior of large-scale circulation features. Every token must appear in the final output (forecasts are spatially dense), but the \emph{intermediate representations} in the middle layers can be approximated by processing only a subset of tokens and broadcasting their updates back via a residual delta.

\textbf{Three-stage block partition.}
Given $N$ transformer blocks, Sparse-Reslim partitions them as:
\begin{equation}
    \underbrace{\text{Block}_{1},\ldots,\text{Block}_{n_1}}_{\text{Dense Early}},\quad
    \underbrace{\text{Block}_{n_1+1},\ldots,\text{Block}_{n_1+n_2}}_{\text{Sparse Middle}},\quad
    \underbrace{\text{Block}_{n_1+n_2+1},\ldots,\text{Block}_{N}}_{\text{Dense Late}},
    \label{eq:partition}
\end{equation}
where $n_1 + n_2 + n_3 = N$. The dense early stage ($n_1$ blocks) builds rich initial representations through full token interaction. The sparse middle stage ($n_2$ blocks) processes only a subset of tokens, yielding the bulk of the speedup. The dense late stage ($n_3$ blocks) refines all tokens before the prediction head, restoring spatial coherence.

\textbf{Random token selection.}
At the boundary between the dense early and sparse middle stages, we randomly select $K = \lfloor L \cdot r \rfloor$ tokens per sample, where $r \in (0,1]$ is the \emph{keep ratio}. Let $\mathcal{I} = \{i_1,\ldots,i_K\} \subset \{1,\ldots,L\}$ be the index set, drawn uniformly without replacement and sorted to maintain spatial ordering. Random selection, rather than attention-score-based selection, has three advantages: (1)~it requires no scoring overhead, (2)~it acts as a stochastic regularizer during training (analogous to dropout on the token dimension), and (3)~it avoids systematic bias that could cause certain geographic regions to be consistently underrepresented.

\textbf{Residual delta formulation.}
Let $\mathbf{x} \in \mathbb{R}^{B \times L \times D}$ denote the full token sequence after the dense early stage. We save a copy $\mathbf{x}_0 = \mathbf{x}$, then gather the selected tokens: $\mathbf{h}_0 = \text{Gather}(\mathbf{x},\;\mathcal{I}) \in \mathbb{R}^{B \times K \times D}.$
These $K$ tokens pass through the $n_2$ sparse middle blocks:
\begin{equation}
    \mathbf{h}_{n_2} = \text{Block}_{n_1+n_2}\!\circ\cdots\circ\text{Block}_{n_1+1}(\mathbf{h}_0).
    \label{eq:sparse_forward}
\end{equation}
The residual delta is then computed and scattered back:
\begin{equation}
    \boldsymbol{\Delta} = \mathbf{h}_{n_2} - \mathbf{h}_0, \qquad
    \mathbf{x}' = \mathbf{x}_0 + \text{Scatter}(\boldsymbol{\Delta},\;\mathcal{I}),
    \label{eq:scatter}
\end{equation}
where $\text{Scatter}$ places the $K$-dimensional deltas into an $L$-dimensional zero tensor at the selected indices. For the $L{-}K$ unselected tokens, $\boldsymbol{\Delta}_i = 0$, so they retain their pre-routing representation $\mathbf{x}_{0,i}$ exactly, forming an implicit identity skip connection. This is a key departure from Sprint~\cite{park2025sprint}, which replaces dropped tokens with learnable mask embeddings and requires an extra linear projection to fuse them back. The mask embeddings never pass through the middle transformer blocks, so they carry a fundamentally different representation distribution than the processed tokens; the fusion layer must learn to compensate for this mismatch. Our formulation sidesteps the issue entirely: unselected tokens retain their actual dense-early representations, the delta scatter is additive and parameter-free, and no fusion layer is needed. We verify this advantage empirically in Table~\ref{tab:selection}, where even within our framework, deterministic selection strategies underperform random selection, confirming that the stochastic regularization from our zero-parameter design is a feature, not a compromise.

\textbf{Computational savings.}
In the sparse middle stage, self-attention operates on $K$ tokens instead of $L$, reducing cost from $O(n_2 L^2 D)$ to $O(n_2 r^2 L^2 D)$. With the default $r{=}0.25$ and $n_1{=}n_3{=}2$, the theoretical FLOPs reduction for the transformer blocks is:
\begin{equation}
    \text{Speedup} \approx \frac{N}{n_1 + n_2 r^2 + n_3} = \frac{12}{2 + 8\cdot 0.0625 + 2} = \frac{12}{4.5} \approx 2.67\times.
    \label{eq:speedup}
\end{equation}
In practice, we observe ${\sim}2.46\times$ speedup on AMD MI250X GPUs, with the gap from theory explained by the overhead of \texttt{gather}/\texttt{scatter} operations and non-transformer components (patch embedding, prediction head, CNN skip path).

\subsection{Integration with Deterministic Forecasting}
\label{sec:deterministic}

For Res-Slim-ViT, integration is straightforward: each block contains only self-attention and MLP with pre-norm residual connections. The sparse middle blocks process only $K$ tokens through the standard block:
\begin{equation}
    \mathbf{h} \leftarrow \mathbf{h} + \text{Self-Attn}(\text{LN}(\mathbf{h})), \quad
    \mathbf{h} \leftarrow \mathbf{h} + \text{MLP}(\text{LN}(\mathbf{h})),
\end{equation}
with attention cost $O(K^2 D)$ per block. All other components (patch embedding, variable aggregation, positional embedding, MLP head, CNN skip path, and unpatchify) remain unchanged and operate on the full $L$ tokens. The complete procedure is given in Algorithm~\ref{alg:sprint_det}.

\begin{algorithm}[t]
\caption{Sparse-Reslim for Deterministic Forecasting (Res-Slim-ViT)}
\scriptsize
\label{alg:sprint_det}
\begin{algorithmic}[1]
\Require Input $\mathbf{X}\in\mathbb{R}^{B\times (H\cdot|V_{\text{in}}|)\times h\times w}$, blocks $\{\text{Block}_i\}_{i=1}^N$, keep ratio $r$, block split $(n_1,n_2,n_3)$
\Ensure Forecast $\hat{Y}\in\mathbb{R}^{B\times|V_{\text{out}}|\times h\times w}$
\State $\mathbf{s} \gets \text{CNN}_{\text{skip}}(\mathbf{X})$ \Comment{Parallel CNN residual path}
\State $\mathbf{Z} \gets \text{VarAgg}(\text{PatchEmbed}(\mathbf{X})) + \mathbf{p}_{\text{pos}} + \mathbf{p}_{\text{res}}$ \Comment{$\mathbf{Z}\in\mathbb{R}^{B\times L\times D}$}
\Statex \textcolor{gray}{\textit{\% Stage 1: Dense early (all $L$ tokens)}}
\For{$i = 1$ to $n_1$}
    \State $\mathbf{Z} \gets \text{Block}_i(\mathbf{Z})$
\EndFor
\Statex \textcolor{gray}{\textit{\% Stage 2: Sparse middle ($K$ random tokens)}}
\State $K \gets \lfloor L \cdot r \rfloor$;\quad $\mathcal{I} \gets \texttt{sort}(\texttt{randperm}(L)[\,:K\,])$
\State $\mathbf{x}_0 \gets \mathbf{Z}.\texttt{clone()}$;\quad $\mathbf{h}_0 \gets \texttt{gather}(\mathbf{Z}, \mathcal{I})$;\quad $\mathbf{h} \gets \mathbf{h}_0.\texttt{clone()}$
\For{$i = n_1{+}1$ to $n_1{+}n_2$}
    \State $\mathbf{h} \gets \text{Block}_i(\mathbf{h})$ \Comment{Self-Attn cost: $O(K^2 D)$}
\EndFor
\State $\boldsymbol{\Delta} \gets \mathbf{h} - \mathbf{h}_0$;\quad $\mathbf{Z} \gets \mathbf{x}_0 + \texttt{scatter}(\mathbf{0}, \mathcal{I}, \boldsymbol{\Delta})$
\Statex \textcolor{gray}{\textit{\% Stage 3: Dense late (all $L$ tokens)}}
\For{$i = n_1{+}n_2{+}1$ to $N$}
    \State $\mathbf{Z} \gets \text{Block}_i(\mathbf{Z})$
\EndFor
\State $\hat{Y} \gets \text{Unpatchify}(\text{MLP}_{\text{head}}(\text{LN}(\mathbf{Z}))) + \mathbf{s}$
\end{algorithmic}
\end{algorithm}

\subsection{Integration with Generative Forecasting}
\label{sec:generative}

The EDM architecture presents a more involved integration because each block contains cross-attention with the conditioning sequence $\mathbf{Z}_c\in\mathbb{R}^{B\times L\times D}$ and noise-conditioned FiLM modulation. During the sparse middle stage, we adopt an \emph{asymmetric attention design}:

\textbf{Self-attention} operates on only the $K$ selected tokens ($\text{Q},\text{K},\text{V} \in \mathbb{R}^{K\times D}$), reducing cost from $O(L^2 D)$ to $O(K^2 D)$.

\textbf{Cross-attention} uses the selected tokens as queries but retains the \emph{full} conditioning sequence as keys and values ($\text{Q}\in\mathbb{R}^{K\times D}$, $\text{K},\text{V}\in\mathbb{R}^{L\times D}$), reducing cost from $O(L^2 D)$ to $O(KL\cdot D)$.

This asymmetry is essential. The conditioning fields $\mathbf{Z}_c$ encode the full atmospheric context (all history frames, all input variables), and restricting keys/values to only $K$ positions would lose meteorological information needed for denoising. The noise embedding $\mathbf{e}_\sigma$ is a global vector broadcast to all tokens and needs no modification; FiLM modulation and the $\sigma$-gate operate element-wise and naturally accommodate the reduced sequence length.

The per-block cost in the sparse middle stage becomes:
\begin{equation}
    \underbrace{O(K^2 D)}_{\text{sparse self-attn}} + \underbrace{O(K L D)}_{\text{sparse cross-attn}} + \underbrace{O(K D^2)}_{\text{MLP + FiLM}},
\end{equation}
compared to $O(L^2 D) + O(L^2 D) + O(L D^2)$ in the dense baseline. The complete procedure is given in Algorithm~\ref{alg:sprint_edm}.

\begin{algorithm}[t]
\caption{Sparse-Reslim for Generative Forecasting (EDM)}
\scriptsize
\label{alg:sprint_edm}
\begin{algorithmic}[1]
\Require Noisy target $\mathbf{y}\in\mathbb{R}^{B\times|V_{\text{out}}|\times h\times w}$, conditions $\mathbf{c}\in\mathbb{R}^{B\times H\times|V_{\text{in}}|\times h\times w}$, noise level $\sigma$, blocks $\{\text{Block}_i^{\text{edm}}\}_{i=1}^N$, keep ratio $r$, split $(n_1,n_2,n_3)$
\Ensure Denoised output $D_\theta(\mathbf{y};\sigma,\mathbf{c})$
\State $c_{\text{in}},c_{\text{out}},c_{\text{skip}} \gets \text{EDM\_precond}(\sigma)$
\State $\mathbf{s} \gets \text{CNN}_{\text{skip}}(c_{\text{in}}\cdot\mathbf{y})$ \Comment{Parallel CNN residual path}
\State $\mathbf{Z}_x \gets \text{VarAgg}(\text{PatchEmbed}(c_{\text{in}}\cdot\mathbf{y})) + \mathbf{p}_{\text{pos}} + \mathbf{p}_{\text{res}} + \mathbf{p}_{\text{time}}$
\State $\mathbf{Z}_c \gets \text{TemporalProj}(\text{VarAgg}(\text{PatchEmbed}(\mathbf{c}))) + \mathbf{p}_{\text{pos}} + \mathbf{p}_{\text{res}}$ \Comment{$\mathbf{Z}_c\in\mathbb{R}^{B\times L\times D}$}
\State $\mathbf{e}_\sigma \gets \text{MLP}(\text{SiLU}(\text{MLP}(\ln\sigma/4)))$ \Comment{Noise embedding}
\Statex \textcolor{gray}{\textit{\% Stage 1: Dense early (all $L$ tokens)}}
\For{$i = 1$ to $n_1$}
    \State $\mathbf{Z}_x \gets \text{Block}_i^{\text{edm}}(\mathbf{Z}_x,\;\mathbf{Z}_c,\;\mathbf{e}_\sigma)$
\EndFor
\Statex \textcolor{gray}{\textit{\% Stage 2: Sparse middle ($K$ random tokens)}}
\State $K \gets \lfloor L \cdot r \rfloor$;\quad $\mathcal{I} \gets \texttt{sort}(\texttt{randperm}(L)[\,:K\,])$
\State $\mathbf{x}_0 \gets \mathbf{Z}_x.\texttt{clone()}$;\quad $\mathbf{h}_0 \gets \texttt{gather}(\mathbf{Z}_x, \mathcal{I})$;\quad $\mathbf{h} \gets \mathbf{h}_0.\texttt{clone()}$
\For{$i = n_1{+}1$ to $n_1{+}n_2$}
    \State $\mathbf{h} \gets \text{Block}_i^{\text{edm}}(\mathbf{h},\;\mathbf{Z}_c,\;\mathbf{e}_\sigma)$ \Comment{Self-Attn: $O(K^2)$; Cross-Attn Q=sparse, KV=full}
\EndFor
\State $\boldsymbol{\Delta} \gets \mathbf{h} - \mathbf{h}_0$;\quad $\mathbf{Z}_x \gets \mathbf{x}_0 + \texttt{scatter}(\mathbf{0}, \mathcal{I}, \boldsymbol{\Delta})$
\Statex \textcolor{gray}{\textit{\% Stage 3: Dense late (all $L$ tokens)}}
\For{$i = n_1{+}n_2{+}1$ to $N$}
    \State $\mathbf{Z}_x \gets \text{Block}_i^{\text{edm}}(\mathbf{Z}_x,\;\mathbf{Z}_c,\;\mathbf{e}_\sigma)$
\EndFor
\State $F_\theta \gets \text{Unpatchify}(\text{MLP}_{\text{head}}(\text{LN}(\mathbf{Z}_x))) + \mathbf{s}$
\State $D_\theta \gets c_{\text{skip}}\cdot\mathbf{y} + c_{\text{out}}\cdot F_\theta$
\end{algorithmic}
\end{algorithm}

\textbf{Training and inference.}
A fresh random permutation is drawn at every forward pass, providing stochastic regularization analogous to dropout on the token dimension. The same random selection is used at inference; averaging multiple runs yields implicit ensembling. For the EDM model, routing is applied identically across all denoising steps.

\textbf{Architecture-agnostic design.}
Sparse-Reslim requires no additional learnable parameters. It introduces only three hyperparameters: the keep ratio $r$ and the block split $(n_1, n_3)$ (with $n_2 = N - n_1 - n_3$). The module wraps the existing block loop and is activated by a single configuration flag. When disabled ($r{=}1$), the model reverts exactly to the dense baseline with zero overhead.

\section{Experiments}
\label{sec:experiments}

\begin{figure}[t]
    \centering
    \includegraphics[width=\textwidth]{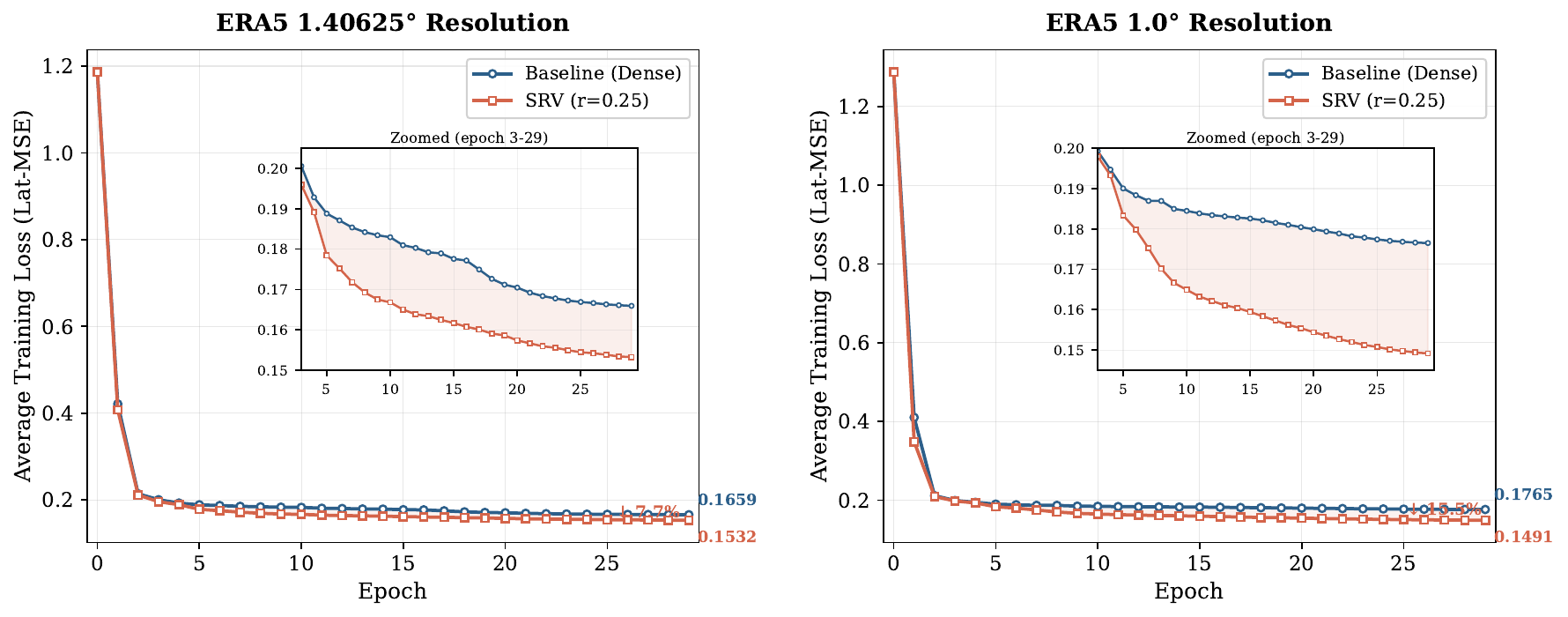}
    \caption{Training convergence for Res-Slim-ViT at 1.40625\textdegree{} and 1.0\textdegree{}. Sparse-Reslim (solid) converges faster than the dense baseline (dashed) in both wall-clock time and training loss, reaching lower final values at both resolutions.}
    \label{fig:convergence}
\end{figure}

\subsection{Experimental Setup}
\label{sec:setup}

\textbf{Dataset.}
We train and evaluate on ERA5 reanalysis~\cite{hersbach2020era5, hersbach2023era5, rasp2020weatherbench, rasp2024weatherbench}, the most widely used benchmark for data-driven weather forecasting. Following the protocol of ClimaX~\cite{nguyen2023climax}, we use 1979--2017 for training, 2018--2019 for validation, and 2020 for testing. The input consists of 40 atmospheric variables: six variables (geopotential, temperature, specific humidity, u-wind, v-wind, relative humidity) at seven pressure levels (50, 250, 500, 600, 700, 850, 925\,hPa), plus four surface and constant fields (2m temperature, 10m u-wind, 10m v-wind, land-sea mask). Evaluation targets four variables standard in NWP verification: geopotential at 500\,hPa (Z500), temperature at 850\,hPa (T850), 2-meter temperature (T2m), and 10-meter zonal wind (U10). Z500 and T850 are upper-air variables used for synoptic-scale assessment; T2m and U10 are surface variables directly relevant to human activities. Unless noted otherwise, experiments use a 5-day (120-hour) forecast lead time.
We run experiments at three spatial resolutions: 1.40625\textdegree{} ($128\times256$ grid, $L{=}8{,}192$ tokens with patch size 2), 1.0\textdegree{} ($181\times360$ grid, $L{=}16{,}200$ tokens with patch size 2), and 0.25\textdegree{} ($720\times1440$ grid, $L{=}16{,}200$ tokens with patch size 8). The 0.25\textdegree{} setting uses the same 1979--2017 training period as the coarser-resolution experiments.

\textbf{Model configurations.}
We evaluate Sparse-Reslim on two backbone architectures:
\emph{Deterministic model (Res-Slim-ViT).} Adapted from ORBIT-2~\cite{wang2025orbit}, this model uses a dual-path architecture with a ViT backbone and a convolutional residual skip path. At 1.40625\textdegree{}, the model has embed\_dim${=}256$, depth${=}12$, four attention heads, totaling 12.18M parameters. At 1.0\textdegree{}, the same architecture on the larger grid yields 14.23M parameters. At 0.25\textdegree{}, we use patch size 8 so that the routed token count matches the 1.0\textdegree{} setting. All deterministic settings use block split $(n_1, n_2, n_3){=}(2, 8, 2)$ with keep ratio $r{=}0.25$.
\emph{Generative model (EDM).} Following the EDM framework~\cite{karras2022elucidating}, this model uses a ViT backbone with FiLM-conditioned noise modulation, $\sigma$-gated cross-attention, and a CNN skip path. The architecture has patch\_size${=}6$, embed\_dim${=}1024$, depth${=}8$, eight attention heads, totaling ${\sim}$180M parameters, with block split $(n_1, n_2, n_3){=}(2, 4, 2)$ and $r{=}0.25$.

\textbf{Implementation details.}
All models are trained with Fully Sharded Data Parallel (FSDP) and activation checkpointing. The deterministic model runs on 2 nodes with 16 AMD MI250X GPUs, using AdamW with learning rate $5{\times}10^{-4}$ (scaled by $\text{num\_gpus}^{0.8}$), cosine schedule with 5 epochs of linear warmup, and batch size 32 (1.40625\textdegree{}) or 16 (1.0\textdegree{} and 0.25\textdegree{}). The generative model is trained on 8 nodes with 64 AMD MI250X GPUs. Both use bfloat16 mixed precision. The latitude-weighted MSE (Eq.~\ref{eq:loss_det}) and EDM denoising loss (Eq.~\ref{eq:loss_edm}) are used for the deterministic and generative models, respectively.

\textbf{Baselines.}
Our primary comparison is the dense baseline, i.e., the identical architecture with sparse routing disabled ($r{=}1$). This isolates the effect of sparse routing from other design choices. We refer to the dense baseline as \textbf{Reslim} and the sparse variant as \textbf{Sparse-Reslim}, trained for the same number of epochs.


\textbf{Evaluation metrics.}
Forecast quality is measured with three complementary metrics: (1)~\textbf{RMSE} (root mean squared error), the standard pointwise accuracy metric; (2)~\textbf{ACC} (anomaly correlation coefficient), which measures correlation between predicted and observed anomalies relative to climatology and is the primary skill score in operational NWP; and (3)~\textbf{SSIM} (structural similarity index), which captures perceptual similarity and penalizes spatial blurring. For efficiency, we report wall-clock training time, throughput (samples per second per GPU), and peak VRAM.

\subsection{Results on Deterministic Forecasting}
\label{sec:det_results}

\begin{figure}[t]
    \centering
    \includegraphics[width=\textwidth]{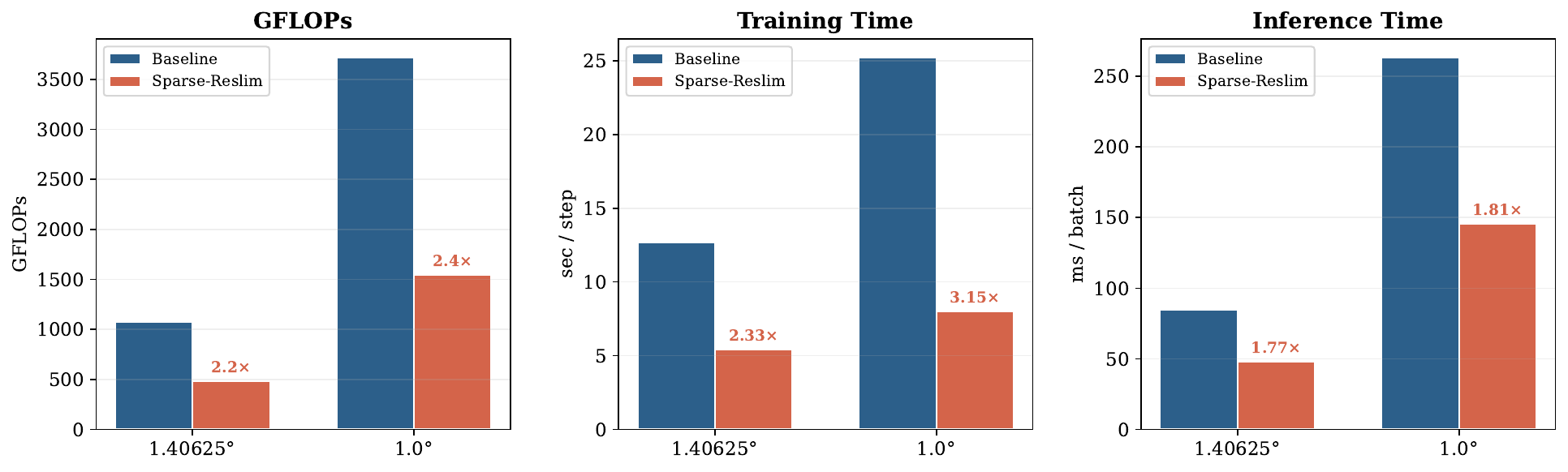}
    \caption{Training efficiency across resolutions for the deterministic model. Sparse-Reslim achieves consistent wall-clock speedup at both resolutions, with larger gains at 1.0\textdegree{} due to the higher token count ($L{=}16{,}200$ vs.\ $8{,}192$).}
    \label{fig:efficiency_res}
\end{figure}

\begin{table}[t]
\centering
\caption{
Forecast quality on the ERA5 test set, reported as single-step, non-autoregressive 120-hour forecasts for the deterministic Res-Slim-ViT. Sparse-Reslim uses $r=0.25$ with block split $(2,8,2)$. The $0.25^\circ$ setting uses the full 1979--2017 training period with patch size 8. \textbf{Bold} = better within each resolution.
}
\label{tab:forecast_quality}
\vspace{4pt}
\scriptsize
\setlength{\tabcolsep}{2.2pt}
\resizebox{\textwidth}{!}{
\begin{tabular}{l l
c c c
c c c
c c c}
\toprule
&
& \multicolumn{3}{c}{$\mathbf{1.40625^\circ}$}
& \multicolumn{3}{c}{$\mathbf{1.0^\circ}$}
& \multicolumn{3}{c}{$\mathbf{0.25^\circ}$} \\
\cmidrule(lr){3-5} \cmidrule(lr){6-8} \cmidrule(lr){9-11}
\textbf{Variable}
& \textbf{Method}
& {RMSE$\downarrow$}
& {ACC$\uparrow$}
& {SSIM$\uparrow$}
& {RMSE$\downarrow$}
& {ACC$\uparrow$}
& {SSIM$\uparrow$}
& {RMSE$\downarrow$}
& {ACC$\uparrow$}
& {SSIM$\uparrow$} \\
\midrule
\multirow{2}{*}{T2m}
& Dense
& 3.18 & 0.798 & 0.858
& 3.19 & 0.797 & 0.857
& 3.31 & 0.775 & 0.865 \\
& Sparse-Reslim
& \textbf{2.84} & \textbf{0.835} & \textbf{0.878}
& \textbf{2.68} & \textbf{0.859} & \textbf{0.894}
& \textbf{2.82} & \textbf{0.846} & \textbf{0.899} \\
\midrule
\multirow{2}{*}{U10}
& Dense
& 3.93 & 0.296 & 0.318
& 4.09 & 0.227 & 0.296
& 4.16 & 0.243 & 0.304 \\
& Sparse-Reslim
& \textbf{3.87} & \textbf{0.341} & \textbf{0.334}
& \textbf{3.83} & \textbf{0.366} & \textbf{0.362}
& \textbf{3.70} & \textbf{0.392} & \textbf{0.374} \\
\midrule
\multirow{2}{*}{Z500}
& Dense
& 872.9 & 0.546 & 0.761
& 868.7 & 0.551 & 0.802
& 1042 & 0.548 & 0.815 \\
& Sparse-Reslim
& \textbf{809.6} & \textbf{0.616} & \textbf{0.794}
& \textbf{768.3} & \textbf{0.651} & \textbf{0.852}
& \textbf{892} & \textbf{0.670} & \textbf{0.865} \\
\midrule
\multirow{2}{*}{T850}
& Dense
& 3.92 & 0.629 & 0.737
& 3.86 & 0.629 & 0.771
& 4.03 & 0.651 & 0.780 \\
& Sparse-Reslim
& \textbf{3.60} & \textbf{0.690} & \textbf{0.763}
& \textbf{3.37} & \textbf{0.726} & \textbf{0.812}
& \textbf{3.45} & \textbf{0.753} & \textbf{0.824} \\
\bottomrule
\end{tabular}
}
\end{table}

Table~\ref{tab:forecast_quality} compares forecast quality between the dense Reslim baseline and Sparse-Reslim across all four target variables and all three resolutions. Sparse-Reslim outperforms the dense baseline on every metric and every variable, often by a substantial margin. At 1.0\textdegree{}, Z500 RMSE drops from 868.7 to 768.3 and ACC rises from 0.551 to 0.651. T2m shows a similar pattern: RMSE improves from 3.19 to 2.68 and ACC from 0.797 to 0.859. The gains are consistent at 1.40625\textdegree{} as well, though slightly smaller in magnitude, which is expected given the lower token count. The 0.25\textdegree{} experiment further confirms the high-resolution behavior under the full 39-year training split: Sparse-Reslim improves all four variables while using the same routed token count as the 1.0\textdegree{} patch-size-2 setting.

Efficiency methods typically aim to \emph{preserve} accuracy, not improve it. Our ablations indicate a dual mechanism: token-level sparsity itself reduces redundant intermediate computation and can improve generalization, while random selection adds a further dropout-like effect without selector parameters or scoring overhead. Figure~\ref{fig:convergence} confirms that Sparse-Reslim converges faster and reaches lower final loss than the dense baseline, ruling out under-training as an explanation. Figure~\ref{fig:efficiency_res} shows that speedup is consistent across resolutions and grows at higher resolutions where the larger token count amplifies quadratic savings.

\begin{table}[t]
\centering
\caption{Peak VRAM per AMD MI250X GCD for the deterministic model. Sparse-Reslim uses $r{=}0.25$ and split $(2,8,2)$.}
\label{tab:peak_vram}
\vspace{4pt}
\scriptsize
\setlength{\tabcolsep}{8pt}
\begin{tabular}{l l c c c}
\toprule
\textbf{Tokens $L$} & \textbf{Setting} & \textbf{Dense (GB)} & \textbf{Sparse (GB)} & \textbf{Reduction} \\
\midrule
8,192  & 1.40625\textdegree{}, p=2 & 26.7 & 12.1 & \textbf{2.20$\times$} \\
16,200 & 1.0\textdegree{}, p=2     & 49.1 & 20.4 & \textbf{2.41$\times$} \\
16,200 & 0.25\textdegree{}, p=8    & 49.3 & 20.6 & \textbf{2.40$\times$} \\
\bottomrule
\end{tabular}
\end{table}

Table~\ref{tab:peak_vram} reports empirical peak VRAM at fixed token counts on 16 AMD MI250X GCDs. Sparse-Reslim reduces peak activation memory by more than $2.2\times$ across all tested settings. The measured reduction is close to the theoretical middle-block estimate $12/(4+8r^2)=2.67\times$ for split $(2,8,2)$ at $r{=}0.25$; the remaining gap comes from full-sequence components such as patch embedding, dense early/late blocks, the prediction head, and the CNN skip path.

\begin{table}[t]
\centering
\caption{Autoregressive rollout stability on the 2020 test year. Both models are trained for 24-hour prediction at 1.40625\textdegree{} and rolled out for 1--10 days. Values are T2m ACC.}
\label{tab:autoregressive_rollout}
\vspace{4pt}
\scriptsize
\setlength{\tabcolsep}{4pt}
\begin{tabular}{l c c c c c c c c c c}
\toprule
\textbf{Method} & \textbf{1} & \textbf{2} & \textbf{3} & \textbf{4} & \textbf{5} & \textbf{6} & \textbf{7} & \textbf{8} & \textbf{9} & \textbf{10} \\
\midrule
Dense & 0.85 & 0.81 & 0.79 & 0.76 & 0.71 & 0.68 & 0.65 & 0.60 & 0.58 & 0.53 \\
Sparse-Reslim & \textbf{0.92} & \textbf{0.89} & \textbf{0.86} & \textbf{0.83} & \textbf{0.80} & \textbf{0.77} & \textbf{0.75} & \textbf{0.73} & \textbf{0.71} & \textbf{0.69} \\
\bottomrule
\end{tabular}
\end{table}

Table~\ref{tab:autoregressive_rollout} addresses the stability of stochastic routing under autoregressive deployment. Sparse-Reslim remains above the dense baseline throughout the 10-day rollout, and the skill decays smoothly rather than showing compounding instability. The result is consistent with the residual identity path: an unselected token is not replaced by noise or a mask embedding, but simply receives zero middle-block delta before being refined by the dense late blocks.

\subsection{Results on Generative Forecasting}
\label{sec:gen_results}

We apply Sparse-Reslim to the EDM-based generative model at 1.0\textdegree{} with the asymmetric attention design described in Sec.~\ref{sec:generative}. The EDM backbone has ${\sim}180$M parameters, depth 8 with split $(2,4,2)$, and uses patch size 6 with embed dim 1024. All metrics are computed on the ensemble mean of 4 samples.

\begin{wrapfigure}{r}{0.6\textwidth}

\centering
\caption{
Forecast quality of the EDM generative model on the ERA5 test set (year 2020, 120-hour lead time) at 1.0\textdegree{}. Sparse-Reslim uses $r{=}0.25$ with block split $(2,4,2)$. \textbf{Bold} = better.
}
\label{tab:edm_quality}
\vspace{4pt}
\scriptsize
\setlength{\tabcolsep}{5pt}
\begin{tabular}{
l l
S[table-format=3.2]
S[table-format=1.3]
}
\toprule
\textbf{Variable}
& \textbf{Method}
& {RMSE$\downarrow$}
& {ACC$\uparrow$} \\
\midrule
\multirow{2}{*}{T2m}
& Dense EDM
& 3.51  & 0.771 \\
& Sparse-Reslim EDM
& \bfseries 3.28  & \bfseries 0.802 \\
\midrule
\multirow{2}{*}{U10}
& Dense EDM
& 4.42  & 0.204 \\
& Sparse-Reslim EDM
& \bfseries 4.17  & \bfseries 0.251 \\
\midrule
\multirow{2}{*}{Z500}
& Dense EDM
& 951.3 & 0.517 \\
& Sparse-Reslim EDM
& \bfseries 889.7 & \bfseries 0.571 \\
\midrule
\multirow{2}{*}{T850}
& Dense EDM
& 4.22  & 0.596 \\
& Sparse-Reslim EDM
& \bfseries 3.95  & \bfseries 0.648 \\
\bottomrule
\end{tabular}
\end{wrapfigure}

Table~\ref{tab:edm_quality} shows that Sparse-Reslim transfers to the generative setting: Z500 RMSE drops from 951.3 to 889.7 and ACC rises from 0.517 to 0.571. T2m and T850 follow the same pattern with RMSE reductions of 6.6\% and 6.4\%, respectively. The improvement margin is smaller than in the deterministic case, which is expected: the EDM backbone uses split $(2,4,2)$ with only 4 sparse middle blocks instead of 8, reducing the computational savings and the strength of stochastic regularization. Sparse-Reslim EDM trains at $1.82\times$ the wall-clock speed of the dense EDM baseline (20.8 vs.\ 11.4 samples/s/GPU), consistent with the smaller $n_2$ and the fact that cross-attention retains full-length conditioning tokens.

\subsection{Ablation Studies}
\label{sec:ablation}

Ablations use the deterministic Res-Slim-ViT at 1.0\textdegree{} unless stated otherwise.

\begin{wrapfigure}{r}{0.52\linewidth}
    \centering
    \includegraphics[width=0.52\textwidth]{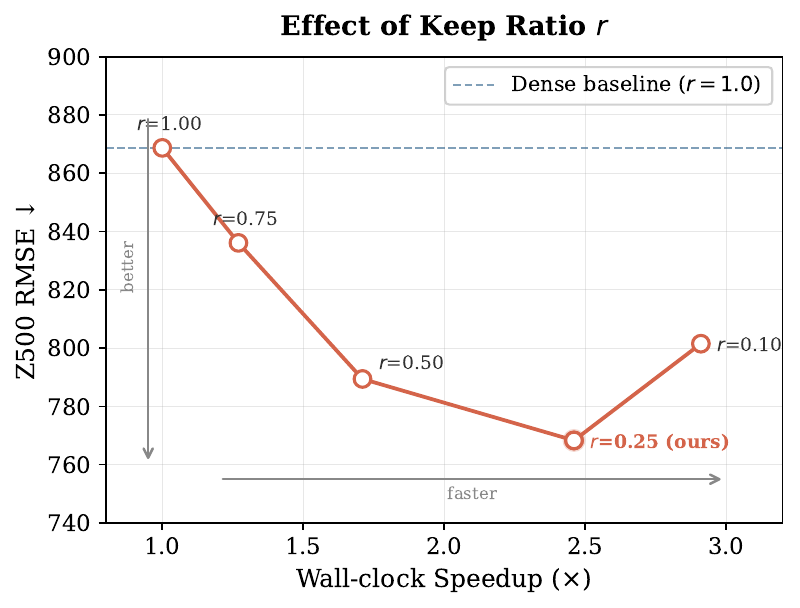}
    \caption{Keep ratio $r$ vs.\ Z500 RMSE and speedup. $r{=}0.25$ is the sweet spot.}
    \label{fig:keep_ratio}
\end{wrapfigure}

\textbf{Effect of keep ratio $r$.}
We vary $r \in \{0.10, 0.25, 0.50, 0.75, 1.0\}$ with a fixed split $(2,8,2)$. Figure~\ref{fig:keep_ratio} plots Z500 RMSE and wall-clock speedup as a function of $r$, where $r{=}1.0$ corresponds to the dense baseline. The accuracy-speedup tradeoff exhibits a clear U-shape: $r{=}0.25$ achieves the lowest RMSE (768.3) and the highest ACC (0.651), outperforming both the dense baseline ($r{=}1.0$, RMSE 868.7) and the more aggressive $r{=}0.10$ (RMSE 801.5). At $r{=}0.10$, the model loses too many tokens and intermediate representations become impoverished; at $r{\geq}0.50$, the stochastic regularization effect diminishes and accuracy degrades toward the dense baseline. The speedup ranges from $1.27\times$ at $r{=}0.75$ to $2.91\times$ at $r{=}0.10$, with $r{=}0.25$ yielding $2.46\times$.

\textbf{Effect of block split $(n_1, n_2, n_3)$.}
Table~\ref{tab:block_split} compares four block partitions for a 12-block model at $r{=}0.25$. The split $(2,8,2)$ produces the best forecast accuracy. Reducing either end to a single dense block, as in $(1,10,1)$, degrades RMSE by 5.7\% despite yielding the highest speedup ($3.12\times$): one dense early block is insufficient for building rich initial token representations, and one dense late block cannot adequately restore spatial coherence. Conversely, allocating more blocks to the dense stages, as in $(4,4,4)$, leaves too few sparse middle blocks for meaningful regularization and limits speedup.


\begin{table}[t]
\centering
\caption{Effect of block split $(n_1,n_2,n_3)$ on Z500 at 1.0\textdegree{}. $r{=}0.25$, $N{=}12$ total blocks.}
\label{tab:block_split}
\vspace{4pt}
\scriptsize
\setlength{\tabcolsep}{10pt}
\begin{tabular}{
  l
  S[table-format=3.1]
  S[table-format=1.3]
  S[table-format=1.2]
}
\toprule
\textbf{Split} $(n_1,n_2,n_3)$
  & {Z500 RMSE$\downarrow$}
  & {Z500 ACC$\uparrow$}
  & {Speedup$\uparrow$} \\
\midrule
$(1,\;10,\;1)$            & 812.4 & 0.609 & 3.12 \\
$\mathbf{(2,\;8,\;2)}$    & \bfseries 768.3 & \bfseries 0.651 & 2.46 \\
$(3,\;6,\;3)$              & 783.6 & 0.638 & 1.78 \\
$(4,\;4,\;4)$              & 831.2 & 0.590 & 1.44 \\
\bottomrule
\end{tabular}
\end{table}

\textbf{Random vs.\ learned token selection.}
A natural question is whether informed token selection can outperform random sampling. Table~\ref{tab:selection} compares three strategies: (1) random selection (our default), (2) top-K by self-attention score from the last dense early block, and (3) a learned difficulty head (a lightweight MLP that predicts per-token importance). Random selection achieves the best Z500 RMSE while adding zero parameters and zero computational overhead. Attention-score-based selection introduces a systematic spatial bias, consistently prioritizing dynamically active regions (jet streams and frontal boundaries) and underrepresenting quiescent tropical areas, which hurts generalization. The learned head partially mitigates this bias but remains deterministic, forgoing the regularization benefit that random selection provides.

\begin{table}[t]
\centering
\caption{Token selection strategy comparison on Z500 at 1.0\textdegree{}. $r{=}0.25$, split $(2,8,2)$.}
\label{tab:selection}
\vspace{4pt}
\scriptsize
\begin{tabular}{
l
S[table-format=3.1]
S[table-format=1.3]
r
r
}
\toprule
\textbf{Selection}
& {Z500 RMSE$\downarrow$}
& {Z500 ACC$\uparrow$}
& {\textbf{Extra params}}
& {\textbf{Overhead (ms/step)}} \\
\midrule
Random (ours) & \bfseries 768.3 & \bfseries 0.651 & 0 & 0 \\
Top-K attn score & 784.1 & 0.636 & 0 & {+14} \\
Learned head & 779.8 & 0.641 & {66K} & {+9} \\
\bottomrule
\end{tabular}
\end{table}

\begin{wrapfigure}{r}{0.5\linewidth}
    \centering
    \includegraphics[width=0.5\textwidth]{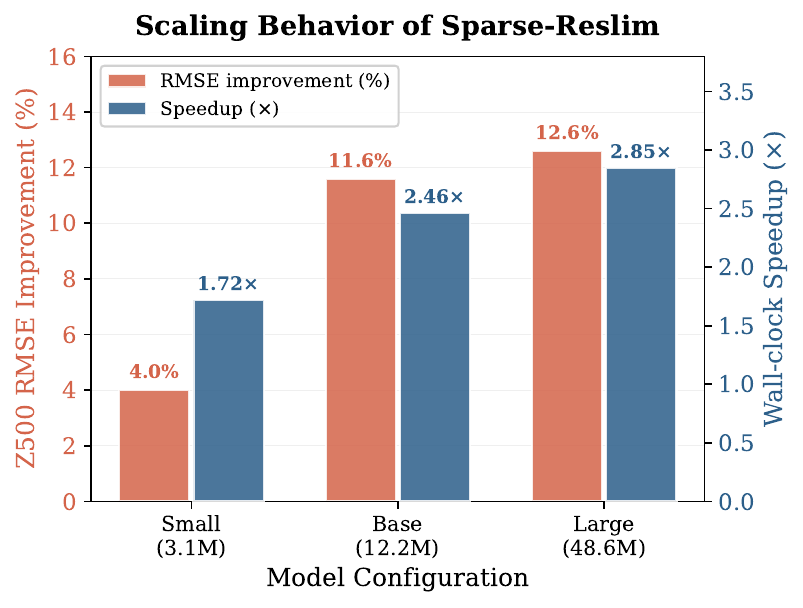}
    \caption{Scaling behavior: both speedup and RMSE improvement grow with model size.}
    \label{fig:scaling}
\end{wrapfigure}

\textbf{Scaling behavior with model size.}
We train three model sizes (Small: embed dim 128, depth 8, 3.1M params; Base: embed dim 256, depth 12, 12.2M; Large: embed dim 512, depth 16, 48.6M) with split $(2,N{-}4,2)$ and $r{=}0.25$. Figure~\ref{fig:scaling} shows that both speedup and accuracy improvement grow with model size. The Large model achieves $2.85\times$ speedup (vs.\ $1.72\times$ for Small) because deeper networks have a larger fraction of their compute in the sparse middle blocks. The RMSE improvement over the dense baseline also grows: $-4.0\%$ for Small, $-11.6\%$ for Base, and $-12.6\%$ for Large, indicating that larger models benefit more from the stochastic regularization provided by random token dropping.

\section{Conclusion}
\label{sec:conclusion}

We introduced Sparse-Reslim, a parameter-free sparse token routing module that partitions ViT blocks into dense early, sparse middle, and dense late stages, processing only a random 25\% of tokens in the dominant middle layers while a residual delta formulation maintains spatially dense outputs. Experiments on ERA5 at three resolutions show ${\sim}2.5\times$ wall-clock training speedup and more than $2.2\times$ peak VRAM reduction alongside improved forecast accuracy: Z500 RMSE drops by 11.6\% at 1.0\textdegree{} relative to the dense baseline, with similar gains for T2m, T850, and U10. We attribute this to a combination of sparse routing and stochastic regularization: deterministic sparse routing already captures much of the gain, while random token selection adds a smaller but consistent benefit without selector parameters or scoring overhead. The method generalizes across two architecturally distinct backbones, a deterministic Res-Slim-ViT and a diffusion-based generative model, remains stable in 10-day autoregressive rollout, and adds no learnable parameters. A natural extension is adaptive token selection that allocates computation to dynamically active regions such as storm fronts and jet streams. 

\noindent\textbf{Acknowledgements.} This manuscript has been authored by UTBattelle, LLC, under contract DE-AC05-00OR22725 with the US Department of Energy (DOE). This research was supported by the ORNL’s AI Initiative sponsored by the Director’s Research and Development Program at ORNL. This work's computing time was supported through the Oak Ridge Leadership Computing Facility.

\clearpage
\bibliographystyle{splncs04}
\bibliography{main}

\clearpage
\appendix
\begin{center}
    \huge \textbf{\texttt{Appendix}}
\end{center}

\section{Implementation Details}
\label{sec:app_implementation}

\subsection{Architecture Specifications}
\label{sec:app_arch}

Tables~\ref{tab:app_arch_det} and~\ref{tab:app_arch_edm} consolidate the full architectural specifications for the deterministic and generative models, respectively.

\begin{table}[]
\centering
\small
\caption{Architecture specification for the deterministic Res-Slim-ViT.}
\label{tab:app_arch_det}
\begin{tabular}{l l}
\toprule
\textbf{Hyperparameter} & \textbf{Value} \\
\midrule
\multicolumn{2}{l}{\textit{Backbone}} \\
\quad Architecture & ViT + CNN skip (Conv $\to$ PixelShuffle $\to$ Conv) \\
\quad Total blocks ($N$) & 12 \\
\quad Embed dimension ($D$) & 256 \\
\quad Attention heads & 4 \\
\quad Head dimension & 64 (${=}256/4$) \\
\quad MLP ratio & 4 \\
\quad MLP hidden dim & 1024 (${=}256\times4$) \\
\quad Patch size & 2 ($1.40625^\circ$, $1.0^\circ$); 8 ($0.25^\circ$) \\
\quad Positional encoding & Sinusoidal (positional + resolution) \\
\midrule
\multicolumn{2}{l}{\textit{Input / Output}} \\
\quad Input variables & 6 vars $\times$ 7 pressure levels + 4 surface = 40 \\
\quad Pressure levels & 50, 250, 500, 600, 700, 850, 925\,hPa \\
\quad Surface / constant fields & T2m, U10, V10, land-sea mask \\
\quad Variable aggregation & Per-variable PatchEmbed + cross-attention (Eq.~\ref{eq:var_agg}) \\
\quad Output variables & 4 (Z500, T850, T2m, U10) \\
\midrule
\multicolumn{2}{l}{\textit{Resolution-Specific}} \\
\quad Grid ($1.40625^\circ$) & $128\times256$, $L{=}8{,}192$ tokens, 12.18M params \\
\quad Grid ($1.0^\circ$) & $181\times360$, $L{=}16{,}200$ tokens, 14.23M params \\
\quad Grid ($0.25^\circ$) & $720\times1440$, $L{=}16{,}200$ tokens with patch size 8 \\
\midrule
\multicolumn{2}{l}{\textit{Sparse-Reslim}} \\
\quad Block split $(n_1, n_2, n_3)$ & $(2,\;8,\;2)$ \\
\quad Keep ratio ($r$) & 0.25 \\
\quad Selected tokens ($K$) & 2,048 ($1.40625^\circ$)\;/\;4,050 ($1.0^\circ$, $0.25^\circ$) \\
\quad Selection strategy & Random (uniform, no replacement, sorted) \\
\bottomrule
\end{tabular}
\end{table}

\begin{table}[t]
\centering
\small
\caption{Architecture specification for the generative EDM model at $1.0^\circ$.}
\label{tab:app_arch_edm}
\begin{tabular}{l l}
\toprule
\textbf{Hyperparameter} & \textbf{Value} \\
\midrule
\multicolumn{2}{l}{\textit{Backbone}} \\
\quad Architecture & ViT + EDM framework~\cite{karras2022elucidating} \\
\quad Total blocks ($N$) & 8 \\
\quad Embed dimension ($D$) & 1,024 \\
\quad Attention heads & 8 \\
\quad Head dimension & 128 (${=}1024/8$) \\
\quad Patch size & 6 \\
\quad Total parameters & ${\sim}$180M \\
\midrule
\multicolumn{2}{l}{\textit{EDM-Specific Components}} \\
\quad Block structure & FiLM $\to$ Self-Attn $\to$ $\sigma$-gated Cross-Attn $\to$ MLP \\
\quad FiLM conditioning & Scale + shift from $\mathbf{e}_\sigma$ \\
\quad Cross-attention gate & $\text{sigmoid}(\mathbf{W}\mathbf{e}_\sigma)$ \\
\quad Noise embedding & $\text{MLP}(\text{SiLU}(\text{MLP}(\ln\sigma/4)))$ \\
\quad CNN skip path & Conv $\to$ PixelShuffle $\to$ Conv \\
\quad Noise distribution & $\ln\sigma \sim \mathcal{N}(P_{\text{mean}},\;P_{\text{std}}^2)$ \\
\quad Loss weighting & $\lambda(\sigma) = (\sigma^2+\sigma_{\text{data}}^2)/(\sigma\cdot\sigma_{\text{data}})^2$ \\
\quad Ensemble size (inference) & 4 \\
\midrule
\multicolumn{2}{l}{\textit{Sparse-Reslim}} \\
\quad Block split $(n_1, n_2, n_3)$ & $(2,\;4,\;2)$ \\
\quad Keep ratio ($r$) & 0.25 \\
\quad Self-attention (sparse stage) & Sparse: $\text{Q},\text{K},\text{V}\in\mathbb{R}^{K\times D}$ \\
\quad Cross-attention (sparse stage) & Asymmetric: $\text{Q}\in\mathbb{R}^{K\times D}$, $\text{K},\text{V}\in\mathbb{R}^{L\times D}$ \\
\bottomrule
\end{tabular}
\end{table}

A key design distinction: the EDM model uses \emph{continuous noise levels} sampled from a log-normal distribution, not discrete diffusion steps.
There is no fixed ``number of training steps'' or DDIM-style sampler.
At inference, an ODE integrator (deterministic or stochastic) produces samples; averaging 4 ensemble members yields the reported metrics.

\subsection{Training Configuration}
\label{sec:app_training}

Table~\ref{tab:app_training} consolidates all training hyperparameters.

\begin{table*}[t]
\centering
\scriptsize
\caption{Full training configuration for the main 120-hour experimental settings.}
\label{tab:app_training}
\setlength{\tabcolsep}{3pt}
\begin{tabular}{l c c c c}
\toprule
\textbf{Setting}
  & \textbf{Det.\,($1.40625^\circ$)}
  & \textbf{Det.\,($1.0^\circ$)}
  & \textbf{Det.\,($0.25^\circ$)}
  & \textbf{EDM\,($1.0^\circ$)} \\
\midrule
\multicolumn{5}{l}{\textit{Optimization}} \\
\quad Optimizer & AdamW & AdamW & AdamW & AdamW \\
\quad Learning rate & $5{\times}10^{-4}$ & $5{\times}10^{-4}$ & $5{\times}10^{-4}$ & {$2{\times}10^{-4}$\,} \\
\quad LR scaling & lr$\,{\times}\,$num\_gpus$^{0.8}$ & lr$\,{\times}\,$num\_gpus$^{0.8}$ & lr$\,{\times}\,$num\_gpus$^{0.8}$ & - \\
\quad LR schedule & Cosine decay & Cosine decay & Cosine decay & {Cosine decay\,} \\
\quad Warmup & 5 epochs (linear) & 5 epochs (linear) & 5 epochs (linear) & {10 epochs\,} \\
\quad Total epochs & {30} & {30} & {30} & {100} \\
\quad Batch size (global) & 32 & 16 & 16 & 16 \\
\quad Weight decay & {$1{\times}10^{-5}$\,} & {$1{\times}10^{-5}$\,} & {$1{\times}10^{-5}$\,} & {$1{\times}10^{-5}$\,} \\
\quad Dropout & 0.0 & 0.0 & 0.0 & 0.0 \\
\midrule
\multicolumn{5}{l}{\textit{Infrastructure}} \\
\quad Precision & bfloat16 & bfloat16 & bfloat16 & bfloat16 \\
\quad Parallelism & FSDP & FSDP & FSDP & FSDP \\
\quad Activation ckpt. & Yes & Yes & Yes & Yes \\
\quad Nodes / GPUs & 2\,/\,16 MI250X & 2\,/\,16 MI250X & 2\,/\,16 MI250X & 8\,/\,64 MI250X \\
\midrule
\multicolumn{5}{l}{\textit{Loss}} \\
\quad Loss function & Lat-weighted MSE & Lat-weighted MSE & Lat-weighted MSE & EDM denoising loss \\
\quad  & (Eq.~\ref{eq:loss_det}) & (Eq.~\ref{eq:loss_det}) & (Eq.~\ref{eq:loss_det}) & (Eq.~\ref{eq:loss_edm}) \\
\midrule
\multicolumn{5}{l}{\textit{Data}} \\
\quad Training years & 1979--2017 & 1979--2017 & 1979--2017 & 1979--2017 \\
\quad Validation years & 2018--2019 & 2018--2019 & 2018--2019 & 2018--2019 \\
\quad Test year & 2020 & 2020 & 2020 & 2020 \\
\quad Lead time & 120\,h & 120\,h & 120\,h & 120\,h \\
\bottomrule
\end{tabular}
\end{table*}

All models are trained on the ORNL Frontier supercomputer using AMD MI250X GPUs (64\,GB HBM2e each).
Fully Sharded Data Parallel (FSDP) distributes model state and gradients across GPUs, and activation checkpointing reduces peak memory at the cost of recomputation during the backward pass.
Both models use bfloat16 mixed precision.
The dense baseline and Sparse-Reslim are trained for the same number of epochs to ensure a fair comparison.

\subsection{Scaling Experiment Configurations}
\label{sec:app_scaling}

Table~\ref{tab:app_scaling} details the three model sizes used in the scaling experiment (Section~\ref{sec:ablation}, Figure~\ref{fig:scaling}).
The block split follows the $(2,\;N{-}4,\;2)$ pattern, so larger models have a greater fraction of their blocks in the sparse middle stage.

\begin{table}[t]
\centering
\small
\caption{Model configurations for the scaling experiment (deterministic Res-Slim-ViT, $1.0^\circ$, $r{=}0.25$).}
\label{tab:app_scaling}
\begin{tabular}{l c c c}
\toprule
 & \textbf{Small} & \textbf{Base} & \textbf{Large} \\
\midrule
Embed dimension & 128 & 256 & 512 \\
Depth ($N$) & 8 & 12 & 16 \\
Parameters & 3.1M & 12.2M & 48.6M \\
Block split $(n_1, n_2, n_3)$ & $(2,\;4,\;2)$ & $(2,\;8,\;2)$ & $(2,\;12,\;2)$ \\
Sparse fraction ($n_2/N$) & 50\% & 67\% & 75\% \\
Wall-clock speedup & $1.72\times$ & $2.46\times$ & $2.85\times$ \\
Z500 RMSE $\Delta$ & $-$4.0\% & $-$11.6\% & $-$12.6\% \\
\bottomrule
\end{tabular}
\end{table}

\subsection{Gather / Scatter Implementation}
\label{sec:app_gather_scatter}

The token selection and residual delta reconstruction are implemented using standard PyTorch indexing.
Indices are sorted after random sampling to preserve spatial ordering, as specified in Algorithm~\ref{alg:sprint_det} (line~6).

\begin{small}
\begin{verbatim}
# Gather: select r*L tokens (sorted for spatial ordering)
K = int(r * L)
I = torch.randperm(L, device=x.device)[:K].sort().values
x0 = x.clone()                   # save pre-routing state
h0 = x[:, I, :]                  # (B, K, D) — gather
h  = h0.clone()

# ... sparse middle blocks operate on h (B, K, D) ...

# Scatter: residual delta formulation
delta = h - h0                    # (B, K, D) — change only
x_out = x0                       # start from saved state
x_out[:, I, :] += delta          # additive scatter
\end{verbatim}
\end{small}

A new random permutation is  drawn at every forward pass during training, providing stochastic regularization.
At inference, the same random selection mechanism is used; averaging multiple forward passes with different seeds yields implicit ensembling (Section~\ref{sec:sparse_routing}).
The gather/scatter overhead is negligible because both operations are simple indexing without additional computation.

\section{Additional Experiments}
\label{sec:app_additional_experiments}

\subsection{Additional Variables and Backbones}
\label{sec:app_extra_variables_backbones}

Table~\ref{tab:app_precipitation} evaluates total precipitation, a more localized and intermittent variable than the four main targets. Sparse-Reslim remains better than the dense baseline, indicating that the sparse routing gain is not limited to smooth upper-air fields.

\begin{table}[t]
\centering
\small
\caption{Total precipitation at 1.40625\textdegree{} on the 2020 test year. RMSE is daily-aggregated mm/day.}
\label{tab:app_precipitation}
\vspace{4pt}
\begin{tabular}{l c c}
\toprule
\textbf{Variable} & \textbf{Dense RMSE / ACC} & \textbf{Sparse-Reslim RMSE / ACC} \\
\midrule
Precipitation & 2.28 / 0.268 & \textbf{2.13 / 0.307} \\
\bottomrule
\end{tabular}
\end{table}

Table~\ref{tab:app_climax} removes the CNN-skip confound by integrating Sparse-Reslim into a ClimaX-style pure ViT backbone with no convolutional residual path. The sparse variant improves all evaluated variables, supporting the claim that Sparse-Reslim functions as a drop-in component beyond the Res-Slim-ViT architecture.

\begin{table}[t]
\centering
\small
\caption{Sparse-Reslim as a drop-in component for a ClimaX-style pure ViT at 1.0\textdegree{}. Values are RMSE / ACC on the 2020 test year.}
\label{tab:app_climax}
\vspace{4pt}
\begin{tabular}{l c c}
\toprule
\textbf{Variable} & \textbf{ClimaX-style ViT} & \textbf{+ Sparse-Reslim} \\
\midrule
T2m  & 3.21 / 0.78 & \textbf{2.92 / 0.83} \\
U10  & 3.98 / 0.22 & \textbf{3.70 / 0.34} \\
T850 & 4.04 / 0.63 & \textbf{3.51 / 0.70} \\
\bottomrule
\end{tabular}
\end{table}

\subsection{Mechanism Ablations}
\label{sec:app_mechanism_ablations}

Table~\ref{tab:app_regularization_decomposition} separates generic stochastic regularization from token-level sparse routing. Stochastic depth improves the dense baseline, and deterministic Top-K sparse routing improves further, showing that sparsity itself accounts for a large part of the gain. Random Sparse-Reslim remains best while adding no selector parameters or scoring overhead.

\begin{table*}[t]
\centering
\small
\caption{Decomposing efficiency and regularization at 1.40625\textdegree{}. Values are RMSE / ACC on the 2020 test year.}
\label{tab:app_regularization_decomposition}
\vspace{4pt}
\setlength{\tabcolsep}{5pt}
\begin{tabular}{l c c c c}
\toprule
\textbf{Variable} & \textbf{Dense} & \textbf{+ StochDepth} & \textbf{+ Top-K Sparse} & \textbf{Sparse-Reslim} \\
\midrule
T2m  & 3.18 / 0.798  & 3.02 / 0.815 & 2.95 / 0.821 & \textbf{2.84 / 0.835} \\
U10  & 3.93 / 0.296  & 3.91 / 0.318 & 3.90 / 0.325 & \textbf{3.87 / 0.341} \\
Z500 & 872.9 / 0.546 & 842.3 / 0.578 & 821.5 / 0.598 & \textbf{809.6 / 0.616} \\
T850 & 3.92 / 0.629  & 3.77 / 0.662 & 3.71 / 0.673 & \textbf{3.60 / 0.690} \\
\bottomrule
\end{tabular}
\end{table*}

We also evaluated two parameter-free content-aware token selectors at 1.40625\textdegree{}: a gradient-based sampler using a Laplacian score and an activation-norm sampler from the first dense block. As shown in Table~\ref{tab:app_alt_selection}, neither matches random selection, which avoids persistent geographic bias while requiring no scoring pass.

\begin{table}[t]
\centering
\small
\caption{Additional token-selection strategies on Z500 at 1.40625\textdegree{}.}
\label{tab:app_alt_selection}
\vspace{4pt}
\begin{tabular}{l c c c c}
\toprule
\textbf{Selection} & \textbf{RMSE$\downarrow$} & \textbf{ACC$\uparrow$} & \textbf{Extra params} & \textbf{Overhead} \\
\midrule
Random (ours) & \textbf{809.6} & \textbf{0.616} & 0 & 0 \\
Gradient (Laplacian) & 818.3 & 0.607 & 0 & +6\,ms \\
Activation norm & 822.1 & 0.601 & 0 & +8\,ms \\
\bottomrule
\end{tabular}
\end{table}

\section{Limitations and Future Work}
\label{sec:app_limitations}

\textbf{Data and evaluation scope.}
This work focuses on ERA5 reanalysis~\cite{hersbach2020era5}, a controlled gridded approximation of atmospheric state rather than raw observational data.
Performance on operational forecast systems with irregular observation networks (e.g., radiosondes, satellite swaths) remains to be validated.
Our main forecast-quality table focuses on single-step 120-hour prediction, and our autoregressive analysis is limited to T2m over 1--10 rollout days.
A broader rollout study across all variables and both deterministic and generative models would further clarify long-horizon deployment behavior.

\textbf{Resolution and variable gap.}
Our experiments include a 0.25\textdegree{} high-resolution setting, but the output space remains limited to a small set of target variables.
State-of-the-art systems such as GraphCast~\cite{lam2023learning} and Pangu-Weather~\cite{bi2023accurate} operate at $0.25^\circ$ with 200+ variables.
Extending Sparse-Reslim to this broader variable set, especially humidity and precipitation across more pressure levels, remains an important direction.

\textbf{Adaptive token selection.}
Random token selection is optimal in our experiments (Table~\ref{tab:selection}), but this may not hold at higher resolutions where the information density is more spatially heterogeneous.
Adaptive strategies that allocate computation to dynamically active regions (storm fronts, jet streams) --- noted as a natural extension in Section~\ref{sec:conclusion} --- deserve further investigation.

\textbf{Subseasonal-to-seasonal forecasting.}
Our evaluation focuses on medium-range forecasting (1--10 day lead times).
Subseasonal-to-seasonal (S2S) forecasting at 2--6 week horizons may require different sparsity schedules, since the relevant spatial scales and predictability sources (e.g., tropical modes, stratospheric coupling) differ substantially.

\textbf{Broader efficiency and adaptation context.} Sparse-Reslim targets train\\ ing-time efficiency for a fixed forecasting task, but a complementary challenge is efficiently adapting a pretrained model when the target domain, resolution, or observation network shifts after deployment. Parameter-efficient and prompt-based adaptation~\cite{xiao2025prompt,zhang2025otvp,zhang2025dpcore, xiao2026not}, including in black-box or model-as-a-service settings~\cite{zhang2025prime,zhang2026adapting}, offers lightweight alternatives to full fine-tuning for this purpose. Feedback-guided generative modeling~\cite{wang2025doctor} similarly suggests a route toward improving the physical plausibility of generative weather forecasts. Combining Sparse-Reslim with such adaptation-time techniques is a promising direction for future work.

\end{document}